\documentclass[smallabstract,smallcaptions]{dccpaper}

\usepackage{epsfig}
\usepackage{amsmath}
\usepackage{amssymb}
\usepackage{color}
\usepackage{url}
\usepackage{hyperref}
\usepackage{multirow}
\newlength{\figurewidth}
\newlength{\smallfigurewidth}
\usepackage{subcaption}

\begin{document}

\title
{\large
\textbf{DeepZip: Lossless Data Compression using \\Recurrent Neural Networks}
}

\author{
Mohit Goyal$^{\dag,\gamma}$, Kedar Tatwawadi$^{\ast}$, Shubham Chandak$^{\ast}$ and Idoia Ochoa$^{\gamma}$\\[0.5em]
$^{\dag}$Department of Electrical Engineering, Indian Institute of Technology Delhi, India\\
$^{\ast}$Department of Electrical Engineering, Stanford University, CA, USA\\
$^{\gamma}$Electrical and Computer Engineering, University of Illinois, Urbana, IL, USA\\
\url{goyal.mohit999@gmail.com}, \url{kedart@stanford.edu}\\
}


\maketitle
\thispagestyle{empty}

\begin{abstract}
Sequential data is being generated at an unprecedented pace in various forms, including text and genomic data. This creates the need for efficient compression mechanisms to enable better storage, transmission and processing of such data. To solve this problem, many of the existing compressors attempt to learn models for the data and perform prediction-based compression. Since neural networks are known as universal function approximators with the capability to learn arbitrarily complex mappings, and in practice show excellent performance in prediction tasks, we explore and devise methods to compress sequential data using neural network predictors. We combine recurrent neural network predictors with an arithmetic coder and losslessly compress a variety of synthetic, text and genomic datasets. The proposed compressor outperforms Gzip on the real datasets and achieves near-optimal compression for the synthetic datasets. The results also help understand why and where neural networks are good alternatives for traditional finite context models.\\
The code and data are available at \url{https://github.com/mohit1997/DeepZip}.
\end{abstract}

\Section{Introduction}
There has been a tremendous surge in the amount of data generated in the past years. Along with image and textual data, new types of data such as genomic, 3D VR, and point cloud data are being generated at a rapid pace. A lot of human effort is spent in analyzing the statistics of these new data formats for designing good compressors. From information theory, we understand that a good predictor naturally leads to good compression. In the recent past, recurrent neural network (RNN) based models have proved  extremely effective in natural language processing tasks such as language translation, semantic parsing and more specifically, in the task of language modeling, which includes predicting the next symbol/character in a sequence \cite{rnn_karpathy}. This raises a natural question: \emph{Can RNN-based models be utilized for effective lossless compression?} In this work, we propose a neural network based lossless compressor for sequential data, named DeepZip. DeepZip consists of two major blocks: an RNN based probability estimator and an arithmetic coding based encoder \cite{Witten:1987:ACD:214762.214771}. 

 Before describing the compression framework in detail, we take a look at the existing literature on lossless sequence compression. We then assess the performance of DeepZip on synthetic data as well as real textual and genomic datasets. We conclude by discussing some observations and future extensions.

\SubSection{Related Work}


Ever since Shannon introduced information theory \cite{shannon1948} and showed that the \textit{entropy rate} is the fundamental limit on the compression rate for any stationary process, there have been multiple works attempting to achieve this optimum. Perhaps the most common compression tool is Gzip (\url{https://www.gzip.org/}). Gzip is based on LZ77 \cite{lz77} and Huffman coding \cite{huffman}. LZ77 is a universal compressor, i.e., it asymptotically achieves the optimal compression rate for any stationary source, without the knowledge of the source statistics. 
LZ77 works by searching for matching substrings in the text appearing before the current position and storing pointers to the matches. Gzip achieves further improvements by using Huffman coding to compress the pointers and other streams generated by LZ77.

LZMA (\url{https://www.7-zip.org/}) is another popular compressor which combines LZ77 with atithmetic coding (described later in detail). More generally, arithmetic coding is a technique for compressing data streams given a probability model for the sequence. A large class of compressors model the data using a conditional probability distribution and then use arithmetic coding as the entropy coding technique. This class includes context-tree weighting (CTW) \cite{ctw} and PPM \cite{ppm}, which efficiently use a mixture of multiple models to generate their predictions.

There has been some related work in the past on lossless compression using neural networks. \cite{schmidhuber1996sequential} discussed the application of character-level RNN model for text, and observed that it gives competitive compression performance as compared with the existing compressors. However, as vanilla RNNs were used, the performance was not very competitive for complex sources with longer memory. Recently, \cite{mahoney1} introduced a different framework for using neural networks for text compression. An RNN was used as a context mixer for mixing the opinions of multiple compressors, to obtain improved compression performance. This was later improved upon by the CMIX compressor \cite{cmix}, which is based on a similar approach that mixes together more than 2000 models using an LSTM context mixer. However, unlike DeepZip it still requires designing of the individual context based compressors, which can heavily depend on the kind of source being analyzed. More recently there has also been some work on word-based and semantically aware models for text compression \cite{cox2016syntactically}. 

\Section{Methods}

Consider a data stream $S^N = \{S_1,S_2,\ldots, S_N\}$ over an alphabet $\mathcal{S}$ which we want to compress losslessly. We next describe in detail the DeepZip compression framework for such a stream, and the specific models used in the experiments.

\SubSection{Framework Overview}
The compressor framework can be broken down into two blocks:

\textbf{Probability predictor}: For a sequence $S^N = \{S_1,S_2,\ldots, S_N \}$, the probability predictor block estimates the conditional probability distribution of $S_r$ based on the previously $K$ observed symbols, where $K$ is a hyperparameter. This probability estimate $\hat{P}(S_r | S_{r-1},\ldots,S_{r-K})$ is then fed into the the arithmetic encoding block. The probability predictor block is modeled as a neural network based predictor.

\textbf{Arithmetic coder block}: This block can be thought of as a reversible Finite-State-Machine (FSM) which takes in the probability distribution estimate for the next symbol $S_r$, $\hat{P}(S_r|S_{r-1},\ldots,S_{r-K})$, and encodes it into a state. The final state is encoded using bits which form the compressed representation of the sequence. 

\SubSection{Encoding-Decoding Mechanism}

\begin{figure}[htbp]
\begin{center}
\includegraphics[width=0.75\textwidth]{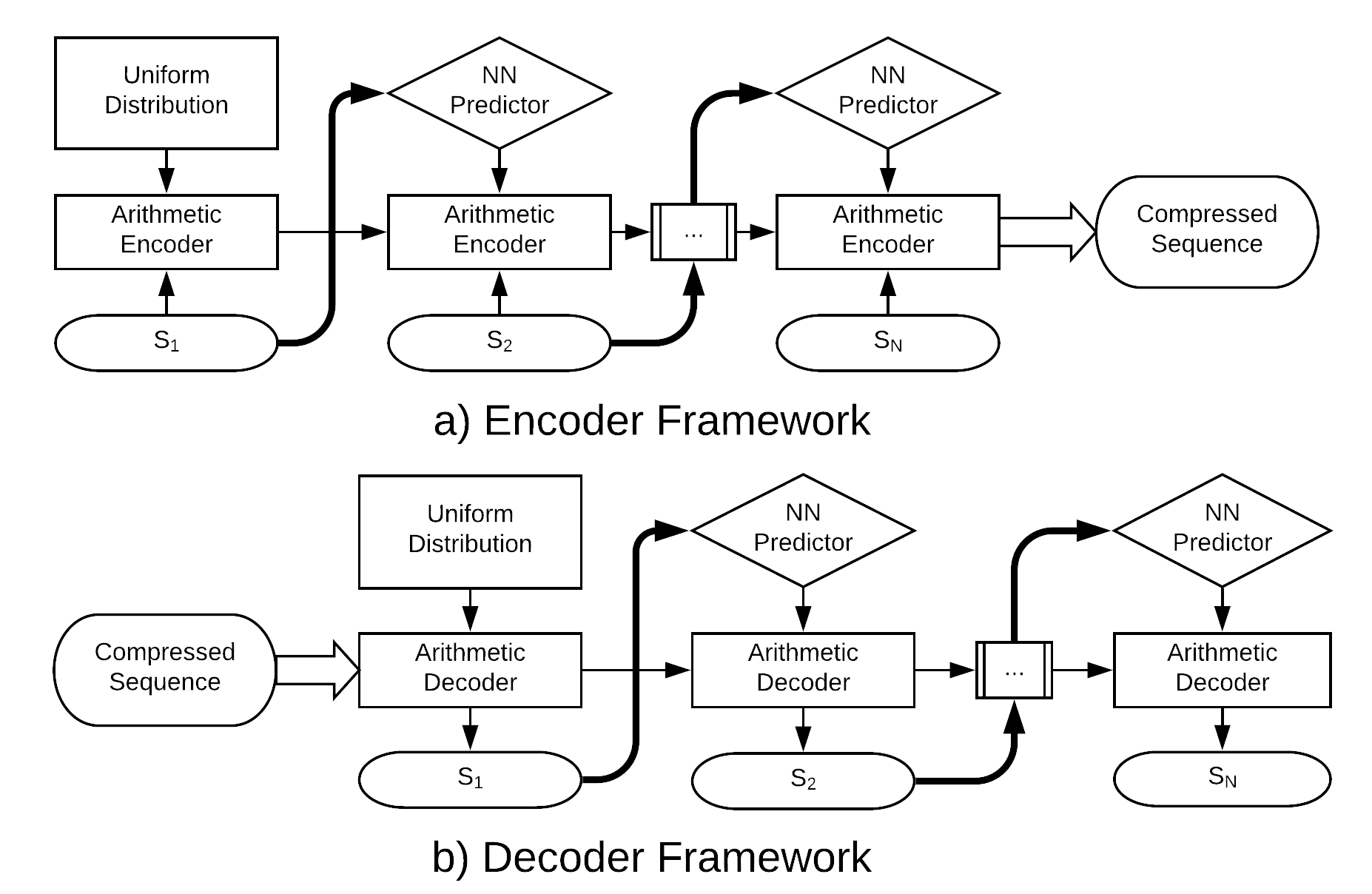}
\end{center}
\vspace{-15pt}
\caption{Encoder-Decoder Framework.}
\label{fig:encoder-decoder}
\end{figure}

The encoding-decoding operations proceed as follows (see Figure \ref{fig:encoder-decoder}):
\vspace{3pt}

1) The neural network model in the probability predictor block is trained on the sequence to be compressed for multiple epochs. Once the training is complete, the model weights are stored, to allow its usage during decompression.
\vspace{3pt}

2) The probability predictor block uses the trained model weights to output a probability distribution over each symbol, which is then used by the arithmetic encoder to perform compression. For the initial $K$ symbols, any prior can be chosen by the arithmetic encoding block. In our framework, we choose a uniform prior, known also to the decoder. Figure \ref{fig:encoder-decoder}a depicts this process for the special case of $K = 1$. 
\vspace{3pt}

3) The operations of the decoder are exactly symmetrical to the encoder, as shown in Figure \ref{fig:encoder-decoder}b. The arithmetic decoder decodes the initial $K$ symbols using a uniform prior distribution, whereas the subsequent symbols are decoded by using the probability distribution provided by the NN-based predictor block. The predictor block utilizes the stored model weights to produce exactly the same probability estimates as the compressor.
\vspace{3pt}

 There are a couple of things which are of utmost important for the correct functionality of DeepZip: Firstly, the probability predictor block needs to be causal, and can have input features based only on the past symbols to ensure all the necessary information is available to the decoder. Secondly, the probability prediction block needs to be perfectly symmetric so as to get back the same probability distribution at the decoder, guaranteeing successful reconstruction of the encoded sequence.



\SubSection{Probability Predictor Block}
We explored several models for the probability predictor block, ranging from fully connected networks to recurrent neural networks such as LSTMs, GRUs, and other variants. This section describes some specific models  and motivation for their use.

    \textbf{Fully connected/dense models (FC)}: A dense or fully connected neural network is a combination of multiple fully connected layers. Mathematically, a dense layer with input $X$ of shape $(batch size, n)$ with $n$ features can be defined as :
    \begin{equation}
        H = \sigma({XW^T + B}),
    \end{equation}
    where $\sigma(x)$ is the activation function, $W$ is the weight matrix, and $B$ is the bias term. In the context of sequence compression, the input for the model would be the previous $K$ symbols and the output would be a multinomial distribution for the next character, $(p_1, p_2, \ldots, p_{|\mathcal{S}|})$, where $|\mathcal{S}|$ is the alphabet size of the sequence. This is obtained by adding a softmax layer at the end, defined as:
    \begin{equation}
    \label{eq:softmax}
        softmax(z)_r = p_r = \frac{e^{z_r}}{\sum_{j=1}^{|\mathcal{S}|} e^{z_j}}
    \end{equation}
    
\begin{figure}[htbp]
            \centering
            \begin{subfigure}[b]{0.35\textwidth}
                \includegraphics[width=\textwidth]{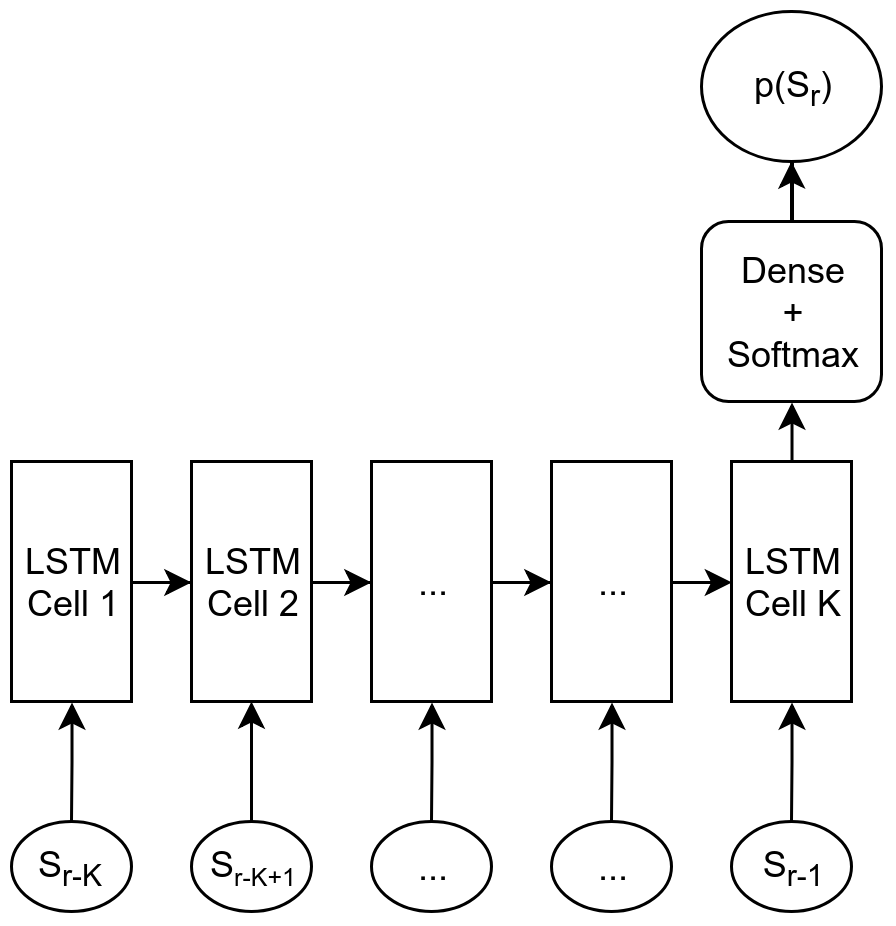}
                \caption{}
                \label{fig:LSTM1}
            \end{subfigure}\hspace{30mm}
            \begin{subfigure}[b]{0.3\textwidth}
                \includegraphics[width=\textwidth]{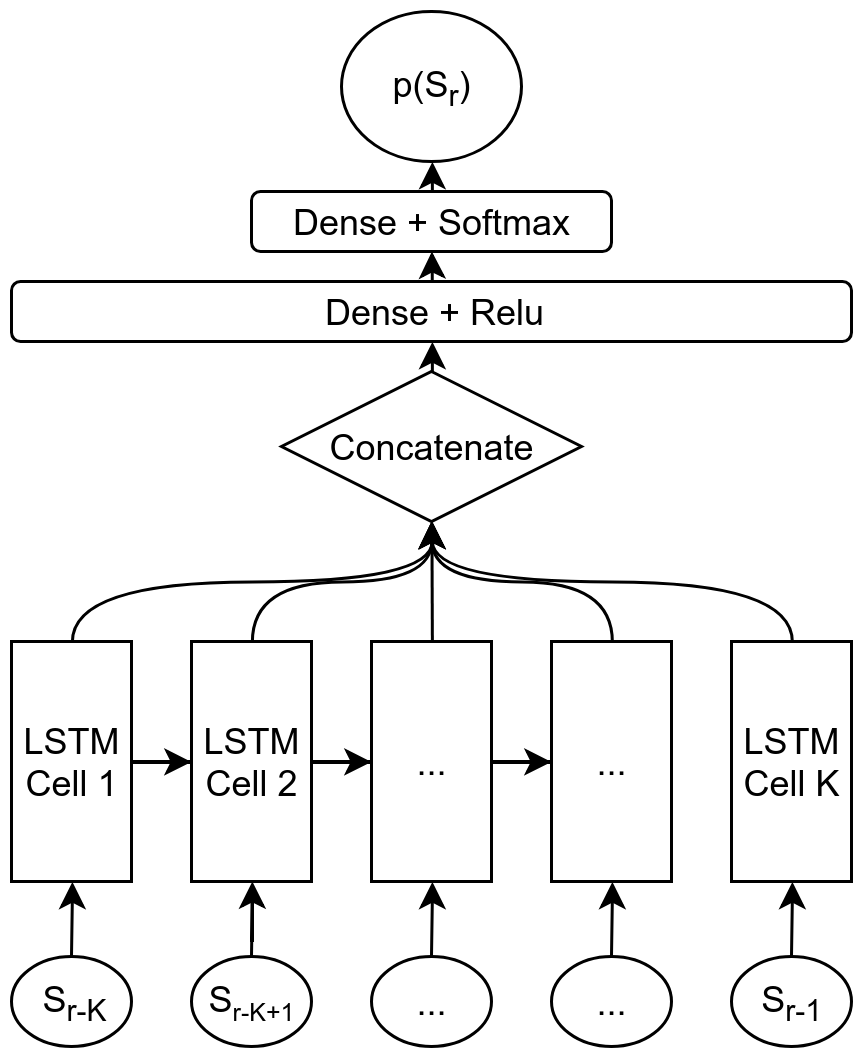}
                \caption{}
                \label{fig:LSTM2}
            \end{subfigure}
            \caption{(a) A multi-input single-output architecture for the probability predictor block. (b) A multi-input multi-output (concatenated) architecture for NN based predictor.}
            \label{fig1}
        \end{figure}

    \textbf{LSTM/GRU single output framework}: LSTM/GRU (Gated Recurrent Units) belong to the class of gated RNNs, and have been used very effectively in the past few years for various natural language processing applications. For every symbol $S_r$, the input consists of the past $K$ symbols $S_{r-K},\ldots,S_{r-1}$. Each of the symbols is an input to a bi-directional GRU cell. Finally, a softmax layer is applied on the final hidden state obtained from the LSTM/GRU cell. The model architecture is illustrated in Figure \ref{fig:LSTM1}. For our experiments, we use a bi-directional variant of the GRU, denoted \textbf{biGRU}.
    

    \textbf{LSTM multi output (concat.) framework:} The LSTM/GRU single output framework suffers from the vanishing gradients issue, making the dependence on farther symbols weak. To alleviate the issue, we consider an LSTM based framework, which includes explicit dependence on all the past $K$ symbols via a fully connected layer on top of LSTM embedding. The architecture is depicted in Figure \ref{fig:LSTM2}. The input consists of past $K$ symbols, which are fed to $K$ LSTM cells. The LSTM cell outputs are concatenated, followed by a dense layer and a subsequent softmax layer which gives the final probability distribution to be used for arithmetic encoding.

\SubSection{Training}

To train the NN predictor based on $K$ previous encountered symbols, with $K$ chosen to be 64, the sequence is divided into overlapping segments of length $K+1$ (shifted by one), where the first $K$ symbols in each segment form the input and the last symbol acts as the output. 
In all of the models described above, the optimizer Adam \cite{adam} is used to minimize categorical cross entropy (default parameters and a batch size fixed at 1024 are used). The model is optimized for a maximum of 10 epochs, where the training is terminated early, if significant improvement is not observed. For every epoch the training data is shuffled, which helps in achieving convergence. We update the model every epoch if there is a decrease in average loss from the previous minima (initially $\infty $). In the LSTM and GRU based models, a cuDNN accelerated implementation \cite{CUDNNLSTM} is used which reduces the training time by approximately 7$\times$.

Note that we do not use cross-validation during the training and in fact attempt to overfit on the training data. This is because the proposed framework stores the model weights as part of the compressed representation and the trained model is used only for prediction on the training data.

\SubSection{Arithmetic coder block}

Arithmetic coding \cite{Witten:1987:ACD:214762.214771} is an entropy coding technique to compress a stream of data given a probability estimate for every symbol, conditioned on the past. Arithmetic coding maintains a range in the interval $[0,1]$. Every stream of symbols uniquely determines a range. This range can be computed sequentially and is directly based on the conditional probability distribution for the next symbol. At the end, the range is encoded into bits, forming the compressed data. The decoder performs the inverse operations, given the probability estimates. We refer the reader to \cite{Witten:1987:ACD:214762.214771} for a detailed description and practical implementation.




To understand how the compressed size for arithmetic coding relates to the categorical cross entropy loss used for training of the prediction models, consider Equation \ref{eq:cat_cross} below which shows the loss function $C(Y, \hat{Y})$, where $Y$ is the one-hot encoded ground truth, $\hat{Y}$ is the predicted probability, $|\mathcal{S}|$ is the alphabet size and $N$ is the sequence length. 
\begin{equation}\label{eq:cat_cross}
    C(Y, \hat{Y}) = \frac{1}{N} \sum_{n=1}^{N}\sum_{k=1}^{|\mathcal{S}|} y_k\log_2 \frac{1}{\hat{y_k}}
\end{equation}
Using the chain rule for probabilities, this expression can be rewritten as shown in Equation \ref{eq:cat_cross_2} where $S^N$ is the sequence and $\hat {p}$ is the joint probability distribution obtained from the predictor block.
\begin{equation}\label{eq:cat_cross_2}
    C(Y, \hat{Y}) = \frac{1}{N} \log_2 \frac{1}{\hat{p}(S^N)}
\end{equation}
Finally, Equation \ref{eq:l_ae} shows that $\bar{L}_{AE}$, the average number of bits used per symbol for arithmetic coding is very close to the loss function from Equation \ref{eq:cat_cross_2}. Thus, categorical cross entropy loss is in fact the optimal loss function to consider while training the models in the DeepZip framework. 
\begin{equation}\label{eq:l_ae}
\frac{1}{N} \log_2 \frac{1}{\hat{p}(S^N)} \leq \bar{L}_{AE} \leq \frac{1}{N} \log_2 \frac{1}{\hat{p}(S^N)} + \frac{2}{N}
\end{equation}



The arithmetic coder in DeepZip is based on an open source Python implementation \cite{arithmetic_library}. 
We achieved significant speedups by parallelizing the encoding and decoding operations. While the computation of predicted probabilities can be easily parallelized during the encoding process, parallelizing the decoding is slightly non-trivial because the computation of probabilities itself depends on the previously decompressed symbols. Thus, we divide the original sequence into $B$ non-overlapping segments during encoding (by default {$B=1000$}). At each step, the probabilities for these segments are computed independently and in parallel by creating a batch of size $B$. This is followed by separate arithmetic coding steps for each segment. 
The decompression process is symmetric to the compression process, and the segments are decoded independently in parallel and concatenated at the end to produce the decoded file.

\Section{Experiments}

We benchmark the performance of our neural network-based compressor DeepZip against Gzip, BSC \cite{bsc}, and some dataset specific compressors like GeCo \cite{geco} (for genomic data) and ZPAQ \cite{zpaq} (for text). For DeepZip we considered the three introduced probability predictor blocks FC, biGRU and LSTM-multi. BSC is a BWT-based compressor which improves over Gzip while being computationally efficient. Several synthetic and real datasets are considered to evaluate the compression that can be achieved with our method and also highlight the advantage which this work provides.\\ 

    \noindent \textbf{- Real datasets:}
    We consider a wide variety of data types including genomic and text data. These datasets were chosen as they benefit in practice from lossless compression. 
    
        \textbf{Human chr1 dataset:} We consider the chromosome 1 DNA refence sequence of the Human Genome Project \cite{hgp}. The alphabet of a DNA sequence typically consists of $\{ A,C,G,T,N\}$, where $\{A,C,G,T\}$ represent the possible nucleotides (bases), and the symbol $N$ represents an unknown nucleotide. Although it is well known that genome sequences have significant repeated regions, state-of-the-art compressors have been unable to capture these repeats, making it a difficult source to compress. 
        
        \textbf{\textit{C.\ Elegans} chr1:} We consider the chromosome 1 of the \emph{C. Elegans} genomic reference sequence for compression, available at \url{ftp://ftp.ensembl.org/pub/release-94/fasta/caenorhabditis_elegans/dna/}.
        
        \textbf{\textit{C.\ Elegans} whole genome:} We also consider the \textit{C.\ Elegans} whole genome sequence for compression, obtained by concatenating its six chromosomes.
        
        \textbf{PhiX virus quality value data:} Along with sequenced nucleotides, raw genomic sequencing data also consists of quality values that represent the probability of error of the obtained nucleotides. We consider 100MB of quality value data for a PhiX virus data, where each symbol takes 4 possible values. Unlike the nucleotide sequences, the quality value sequences are highly compressible since most quality values are the same and correspond to the best quality.
        
        \textbf{text8 dataset:} Along with genomic datasets, we also consider the text8 dataset, which is an ASCII text dataset of size 100MB. The text8 dataset has been widely studied and experimented on in the literature. It can be accessed at \url{http://mattmahoney.net/dc/text8.zip}.\\
        
  \noindent \textbf{- Synthetic datasets:} We generate data from synthetic sources of known entropy rate. Since entropy rate provides a lower bound on the compression ratio, it allows us to gauge the performance of a compression algorithm against this ideal bound. The following sources are considered:
   
        \textbf{Independent and identically distributed (IID):} IID binary data distributed as $Bern(0.1)$ is considered since existing compressors perform fairly good on IID sequences.
        
        \textbf{k-order Markov (XOR):} The Markov-k sources are generated as follows:
        \begin{equation}
            S_{n+1} = S_n + S_{n-k}\ (\text{mod}\ M),
        \end{equation}
        where $M$ is the alphabet size (2 by default). This source is closely related to the lagged Fibonacci pseudorandom generator \cite{laggedfibonacci} and hence is difficult to compress for most traditional compressors, even though the entropy rate for Markov-k sources is in fact 0. We consider $k=\{20,30,40,50\}$ for our experiments.
        
        \textbf{Hidden Markov Model (HMM):} HMM is a statistical Markov model where the system being modeled is assumed to be a Markov process with unobserved hidden state \cite{hmm}. We simulate a HMM source where the hidden state follows the Markov-k sequence described earlier. Specifically, the HMM process is generated as follows:
        \begin{equation}
            S_{n+1} = X_n + X_{n-k} + Z_n\ (\text{mod}\ M). 
        \end{equation}
        Here, the hidden process $H = X_{n-1} + X_{n-k-1}\ (\text{mod}\ M)$ is Markov-k, and $Z_n$ is the added IID noise. We consider $Z_n \sim Bern(0.1)$ and $k=\{20,30,40\}$ for our experiments.

For all our experiments we used a NVIDIA TITAN X GPU (12GB). All the training and encoding-decoding scripts are available at: \url{https://github.com/mohit1997/DeepZip}

\Section{Results and discussion}
        \begin{table}[]
    \begin{tabular}{|c|c|c|c|c|c|c|}
    \hline
    \multirow{2}{*}{\textbf{Dataset}}         &   \multirow{2}{*}{\textbf{Seq. Length}} &\multirow{2}{*}{\textbf{Gzip}} & \multirow{2}{*}{\textbf{BSC}} & \multicolumn{3}{c|}{\textbf{DeepZip}} \\ 
    \cline{5-7}
          &  &  &      & \textbf{FC} & \textbf{biGRU}    & \textbf{LSTM-multi} \\ \hline
    \textit{H. chr1}      & 249M                                 & 60.58      & 50.43          & 49.37    & 48.80          & \textbf{48.56}   \\ \hline
    \textit{C. E. chr1}   & 15M                                  & 4.03       & \textbf{3.49}  & 3.81     & 3.58           & 4.02             \\ \hline
    \textit{C. E. genome} & 100M                                 & 26.97      & 23.38          & 23.41    & \textbf{23.13} & 23.41            \\ \hline
    \textit{text8
    } & 100M                                 & 33.05      & \textbf{20.95} & 25.49    & 23.37          & 26.71            \\ \hline
    \textit{PhiX Quality} & 100M                                 & 6.22       & 4.38           & 4.58     & \textbf{4.35}                 & 4.79             \\ \hline
    \end{tabular}
    \caption{Compression sizes in MB ($10^6$ Bytes) for real datasets. Best results are boldfaced.}
    \label{table:results_real}
    \end{table}
    
    \begin{table}[]
    \centering
    \begin{tabular}{|c|c|c|c|c|c|c|}
    \hline
    \multirow{2}{*}{\textbf{Dataset}}     &  \multicolumn{2}{c|}{\textbf{FC}} & \multicolumn{2}{c|}{\textbf{biGRU}}    & \multicolumn{2}{c|}{\textbf{LSTM-multi}} \\ \cline{2-7}
    & Model & Sequence & Model & Sequence & Model & Sequence\\
    \hline
    \textit{H. chr1}      &  0.39   &  48.98  &  0.17         &  48.62   &  0.62        & 47.95   \\ \hline
    \textit{C. E. chr1}   &  0.39   & 3.42  & 0.17   &  3.40    &   0.62        & 3.98             \\ \hline
    \textit{C. E. genome} &  0.39   & 23.02  & 0.17          &  22.96  &  0.62 &  22.79 \\ \hline
    \textit{text8} & 0.40    & 25.09  & 1.74 & 21.63    & 0.63 & 26.09          \\ \hline
    \textit{PhiX Quality} & 0.39    & 4.19  &    0.17        & 4.18    &  0.62               &    4.18         \\ \hline
    \end{tabular}
    \caption{Breakdown of compression size in MB ($10^6$ Bytes) into model size and size of sequence compressed with arithmetic coding, for DeepZip.}
    \label{table:results_modelsize}
    \end{table}

    \begin{table}[]
    \centering
    \begin{tabular}{|c|c|c|c|c|c|c|}
    \hline
    \multirow{2}{*}{\textbf{Dataset}}         &   \multirow{2}{*}{\textbf{Seq. Length}} &\multirow{2}{*}{\textbf{Gzip}} & \multirow{2}{*}{\textbf{BSC}} & \multicolumn{3}{c|}{\textbf{DeepZip}} \\ 
    \cline{5-7}
          &  &  &      & \textbf{FC} & \textbf{biGRU}    & \textbf{LSTM-multi} \\ \hline
    \textit{IID}     & 10M             & 0.81        & \textbf{0.60} & 0.98          & 0.76          & 1.20             \\ \hline
    \textit{XOR20}   & 10M             & 1.51      & \textbf{0.06}           & 0.40           & 0.18 & 0.63              \\ \hline
    \textit{XOR30}   & 10M             & 1.51       & 1.26         & 0.40         & \textbf{0.18} & 1.87             \\ \hline
    \textit{XOR40}   & 10M             & 1.49       & 1.26         & \textbf{0.40}  & 1.43         & 1.87             \\ \hline
    \textit{XOR50}   & 10M             & 1.48       & 1.26         & 0.40           & \textbf{0.18} & 0.63              \\ \hline
    \textit{HMM20}   & 10M             & 1.49       & 0.87          & 0.98          & \textbf{0.76} & 1.87             \\ \hline
    \textit{HMM30}   & 10M             & 1.49       & 1.26         & 0.98          & \textbf{0.76} & 1.21             \\ \hline
    \textit{HMM40}   & 10M             & 1.49       & 1.26         & \textbf{0.98} & 1.42         & 1.87             \\ \hline
    \end{tabular}
    \caption{Compression sizes in MB ($10^6$ Bytes) for synthetic datasets. Best results are boldfaced.} 
    \label{table:results_synthetic}
    \end{table}

Table \ref{table:results_real} shows the compression results for the real datasets. We compare general purpose compressors Gzip and BSC to the proposed neural network based compressors DeepZip. 
We observe that the proposed compressor outperforms Gzip by about 20\% on text and genomic data. As compared to BSC, DeepZip usually achieves comparable results, with slightly better compression on the \textit{C.\ Elegans} genome. We also observe that for DeepZip, biGRU exhibits in general the best performance.

We also tested some specialized compressors for these datasets. For the human and \textit{C.\ Elegans} genomes, we used GeCo, which achieves 5-10 \% smaller size as compared to DeepZip. Similarly, for text compression, ZPAQ achieves a compressed size of 17.5MB on the text8 dataset \cite{text8results}, which is 25\% lower than that for DeepZip. These results are to be expected, since the specialized compressors typically involve handcrafted contexts and mechanisms which are highly optimized for the dataset statistics. Also, they can improve the compression performance by taking into account that the datasets can in fact be non-stationary. In contrast, the proposed compressor achieves reasonably good results on a wide variety of datasets.

Table \ref{table:results_modelsize} shows the breakdown of size between the model weights and the arithmetic coded stream for the proposed compressor. We observe that the model size contributes significantly to the overall size, especially when the sequence length is small. Currently the model weights are represented as 32 bit floats without further compression. We attempted to use 16 bit floats and TensorFlow Lite \cite{tflite}, but faced stability and compatibility issues. We believe that the model size can be reduced significantly without losing compression performance by using techniques similar to those outlined in \cite{deepcompression}. Furthermore, in some cases, the model can be shared between different sequences, for example when compressing genomes of different individuals which are very similar.

To better understand the ability of the proposed framework, we also experimented with some synthetic data of low Kolmogorov complexity, but which are not compressed well by traditional compressors. 
Table \ref{table:results_synthetic} shows the results for these datasets (IID, XOR and HMM). We see that as we go towards sequences with long-term dependencies, the traditional compressors fail to achieve good compression, only achieving 1 bit per binary symbol. The proposed compressor DeepZip, on the other hand, is able to exploit the structure in these sequences to achieve much better compression. There is still some gap from the entropy of the sequences because of the space needed to store the model and some overhead associated with arithmetic coding. Note that, we observe that in some cases, XOR40 dataset for e.g., the DeepZip performance is significantly dependent upon the training parameters. This can be understood by the fact that the source is pseudo-random, making it difficult for the optimization process to find the appropriate minima. 

Regarding the running time of DeepZip, we observe that for typical datasets of size 10MB, every training epoch requires 1-2 hrs (with a 12GB NVIDIA TITAN X GPU and a batch size of 1024). We typically train every dataset for 3-4 epochs. The encoding/decoding requires performing a single forward pass through the NN-model, and takes approximately 5-10 mins, depending upon the model. 


\Section{Conclusion}
We proposed a neural network prediction based framework for lossless compression of sequential data. The proposed compressor DeepZip achieves improvements over Gzip for a variety of real datasets and achieves near optimal compression for synthetic datasets. Future work involves improving the performance of the compressor, for example by using attention models to improve the prediction and hence the overall compression. We also plan to work on improving compression on non-stationary sources by allowing model weights to be fine-tuned as the sequence is compressed/decompressed, so as to adapt quickly to changing statistics.

Finally, we believe the compression experiments should also help in improving our understanding of the neural network models themselves. The well established information theoretic framework for data compression can be potentially useful for this cause.

\Section{References}
\bibliographystyle{IEEEbib}
\bibliography{refs}



\end{document}